\documentclass[a4paper]{article}

\usepackage{packages/INTERSPEECH2022}
\usepackage{eqparbox}
\usepackage{microtype}
\usepackage{cite}
\usepackage{amsmath,bm}
\usepackage{mathtools,nccmath}
\usepackage{etoolbox,siunitx}
\robustify\bfseries
\usepackage[breaklinks=true,colorlinks,bookmarks=false]{hyperref}
\usepackage{enumitem}
\setlist[itemize]{wide=0pt, leftmargin=10pt, labelwidth=5pt, align=left}

\title{Accelerating Inference and Language Model Fusion of Recurrent Neural Network Transducers via End-to-End 4-bit Quantization}
\name{Andrea Fasoli, Chia-Yu Chen, Mauricio Serrano, Swagath Venkataramani, \\ George Saon, Xiaodong Cui, Brian Kingsbury, Kailash Gopalakrishnan}
\address{IBM Research, USA}
\email{
\{andrea.fasoli,swagath.venkataramani\}@ibm.com\\
\{cchen,mserrano,gsaon,cuix,bedk,kailash\}@us.ibm.com
}
\begin{document}

\DeclarePairedDelimiter{\nint}\lfloor\rceil

\maketitle
\begin{abstract}

We report on aggressive quantization strategies that greatly accelerate inference of Recurrent Neural Network Transducers (RNN-T). We use a 4 bit integer representation for both weights and activations and apply Quantization Aware Training (QAT) to retrain the full model (acoustic encoder and language model) and achieve near-iso-accuracy. We show that customized quantization schemes that are tailored to the local properties of the network are essential to achieve good performance while limiting the computational overhead of QAT.

Density ratio Language Model fusion has shown remarkable accuracy gains on RNN-T workloads but it severely increases the computational cost of inference. We show that our quantization strategies enable using large beam widths for hypothesis search while achieving streaming-compatible runtimes and a full model compression ratio of 7.6$\times$ compared to the full precision model.

Via hardware simulations, we estimate a 3.4$\times$ acceleration from FP16 to INT4 for the end-to-end quantized RNN-T inclusive of LM fusion, resulting in a Real Time Factor (RTF) of 0.06. On the NIST Hub5 2000, Hub5 2001, and RT-03 test sets, we retain most of the gains associated with LM fusion, improving the average WER by $>$1.5\%.

\end{abstract}

\section{Introduction}

End-to-end (E2E) models such as the Recurrent Neural Network Transducer (RNN-T)~\cite{graves2012sequence} map an acoustic feature sequence ${\bf x}$ to a sequence of characters, sub-words, or words ${\bf y}$, directly defining the posterior probability $P({\bf y}|{\bf x})$ without additional conditional independence assumptions. They dramatically simplify both training and decoding pipelines compared to competing approaches for Automatic Speech Recognition (ASR) and demonstrated excellent performance on standard benchmarks~\cite{tuske2020single,baevski2020wav2vec,chung2021w2v,saon2021advancing,guo2021recent}.

RNN-T models have several properties that make them attractive for ASR, such as the lack of a need for pre-existing alignments in training, the incorporation of a recurrent network Language Model (LM) (which is jointly trained with the acoustic encoder), and a computation which is monotonic in time suitable for streaming recognition~\cite{sainath2020streaming,li2020developing}.

In order to boost accuracy, separate LMs, which provide an estimate of $P({\bf y})$, can be trained on larger text-only corpora and combined with the RNN-T outputs using shallow fusion. In particular, Density Ratio fusion~\cite{mcdermott2019density,variani2020hybrid} is an inference-only method which extends popular shallow fusion methods. Yet, shallow fusion can result in a dramatic increase in model size and computational time.

In this work, we examine mixed-precision quantization strategies, primarily at 4 bits, based on Quantization Aware Training (QAT) to alleviate the challenges posed by LM fusion approaches. With QAT, the network is retrained while simulating low-precision operations, allowing it to learn to compensate for the error introduced by the quantization process. The resulting model can be efficiently compressed and accelerated at inference time, ideally with minimal or no loss in accuracy.

Previous works on the acceleration of E2E ASR models for inference tasks demonstrated minimal degradation with 8-bit quantization of the matrix multiplications (both weights and activations), applied to both the acoustic encoder and the LM~\cite{he2019streaming,shangguan2019optimizing,nguyen2020quantization,kim2021integer}. In addition, ultra low precision (1 or 2 bits) representation of the LMs weights alone has been extensively investigated~\cite{ardakani2018learning,liu2018binarized,yu2020neural,xu2021mixed}. To the best of our knowledge, only~\cite{fasoli21lstm} has reported on 4 bit quantization of weights and activations of an E2E ASR model (RNN-T), showing a 1.3\% average WER degradation on the Hub5 2000 test set.

Our main contributions are summarized as follows:
\vspace{-0.05cm}
\begin{itemize}[noitemsep]
    \item We carefully assign quantizers and precisions to both weights and activations of the RNN-T. We stress that while weight quantization is sufficient for model compression, in order to achieve a significant acceleration, quantization of the activations is also a requirement. Our selections are driven by the minimization of WER and by the need to keep the process practical by limiting the retraining time imposed by QAT. Our scheme achieves near-iso-accuracy at 4 bits.
    \item In the context of Density Ratio LM fusion, we fully quantize the separate LMs and demonstrate that the accuracy gains due to fusion translate well to the E2E-quantized RNN-T with quantized additional LMs.
    \item Via hardware (HW) runtime simulations, we show that the workload can be effectively accelerated, enabling hypothesis search at beam width 16 to achieve a Real Time Factor (RTF, defined as processing time over audio duration) competitive with small beam width search using non-quantized models without LM fusion.
\end{itemize}

\section{RNN-T with LM fusion}
\label{sec:models}

\begin{figure}[t]
  \centering
  \includegraphics[width=\linewidth]{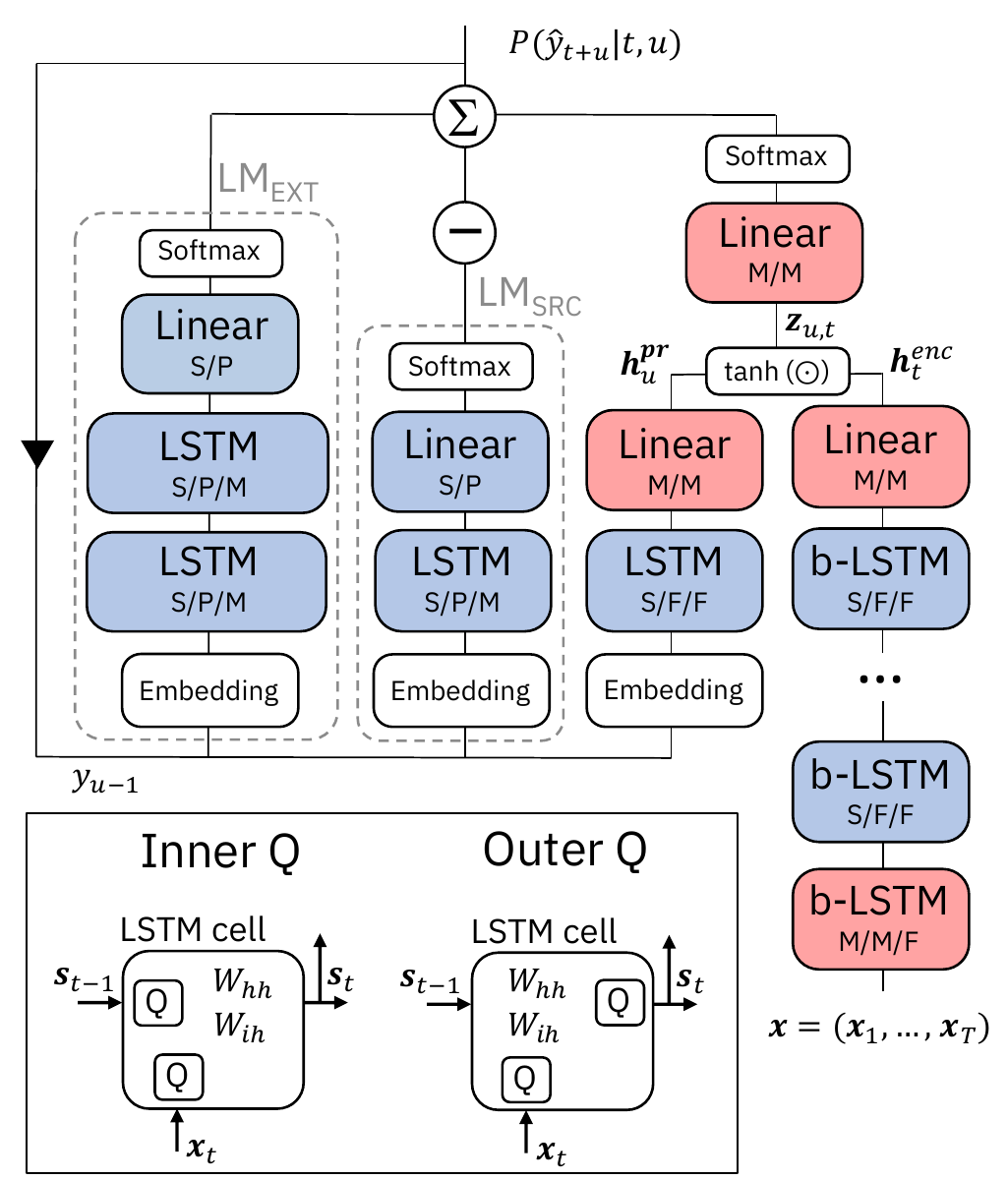}
  \caption{Architecture and main quantization scheme of RNN-T with LMs. Red cell: 8 bits; blue cell: 4 bits. Layers are labelled with their respective quantizers (Linear: weight/activation; LSTM weight/input/hidden state). Abbreviations: F = FIX, M = MAX, P = PACT, S = SAWB, b-LSTM = bidirectional LSTM. Inset: Inner vs. Outer quantization scheme of inputs and hidden states of LSTM cell}
  \label{fig:arch}
  \vspace{-4mm}
\end{figure}

The base RNN-T (Fig.~\ref{fig:arch}) consists of an acoustic encoder, a prediction network, and a joint network. The encoder comprises 6 bi-directional LSTM layers (input size 340 - including 100-dim i-vectors) and a linear layer that generates the latent acoustic representation $h_t^{enc}$, of dimension 256. The prediction network is a single unidirectional LSTM layer (hidden size 768), preceded by a small embedding layer, and followed by a linear layer with output $h_u^{dec}$ also of size 256. Encoder and prediction network outputs are combined multiplicatively in a joint network~\cite{saon2021advancing}, with an additional linear layer and log-Softmax over 46 output characters. In this configuration, the RNN-T comprises 57.2~M parameters (54.6~M in the encoder, 2.6~M in the prediction network). We train for 20 epochs with 64-utterance batches on audio and character-level transcripts from the SWB-1 data collection, augmented with speed and tempo perturbation~\cite{ko2015audio}, SpecAugment~\cite{park2019specaugment}, and Sequence Noise Injection~\cite{saon2019sequence}. We use the AdamW optimizer and the OneCycleLR learning rate scheduler.

For density ratio LM fusion, we train an external LM (LM\textsubscript{EXT}) on the 2000 hours Switchboard + Fisher acoustic transcripts and a source LM (LM\textsubscript{SRC}) on the 300 hours Switchboard transcripts. LM\textsubscript{EXT} consists of an embedding layer, two unidirectional LSTM layers (hidden size = 2048), a bottleneck layer of size 256, and an output layer of size 46 (see Fig.~\ref{fig:arch}). The total number of parameters is 51.0~M, almost 20$\times$ larger than the original prediction network. LM\textsubscript{EXT} is trained with Nesterov SGD for 40 epochs, at a constant learning rate (LR) = 0.03 for 20 epochs, then exponentially decreased by 1/$\sqrt{2}$/epoch. LM\textsubscript{SRC} has the same building blocks and size as the RNN-T prediction network (2.6~M parameters) and is trained for 40 epochs. 

At inference, the symbol predicted at the previous step $y_{u-1}$ is fed back into the prediction network, LM\textsubscript{EXT}, and LM\textsubscript{SRC}, generating 3 prediction scores that are combined as \cite{mcdermott2019density}:
\begin{equation}
    \addtolength{\abovedisplayskip}{-1mm}
    \addtolength{\belowdisplayskip}{-1mm}
    S = \log P(\textbf{y}|\textbf{x}) + \mu \log P_{ext}(\textbf{y}) - \lambda \log P_{src}(\textbf{y}) + \rho |\textbf{y}|
\end{equation}
where $S$ is the combined score, $P$, $P_{ext}$, and $P_{src}$ the probability of the original RNN-T network, LM\textsubscript{EXT}, LM\textsubscript{SRC}, respectively, and $\mu$, $\lambda$, and $\rho$ are the LM weights and character insertion reward, set to $0.7$, $0.5$, $0.2$, respectively.

We use alignment-length synchronous beam search~\cite{saon2020alignment} with variable beam width (1 to 16), corresponding to the maximum number of hypotheses that are retained and evaluated at each iteration of the decoding loop. Decoding is performed on the Hub5 2000, Hub5 2001, and RT-03 test sets and Kaldi scoring is used for measuring WER.

\section{Quantization strategies}\label{sec:quantization_strats}

\subsection{Background}\label{ssec:quantizers}

Quantization is the process of mapping high-precision floating point numbers into a lower bit representation. In particular, low-precision arithmetic, which expresses values as integers multiplied by a scaling factor, is well suited for AI-accelerating HW and enables faster and highly energy efficient execution of Multiply Accumulate (MAC) operations and data transfer. Linear Quantization (LQ) is the most widely adopted approach in QAT, as it allows for a direct implementation of integer arithmetic. Asymmetric LQ computes:

\begin{equation}\label{eq:asymmetric_lq_xint}
    \addtolength{\abovedisplayskip}{-1mm}
    \addtolength{\belowdisplayskip}{-1mm}
    x_{int} = clamp( \nint{\frac{x}{s}} + z; 0; 2^b - 1)
\end{equation}

\begin{equation}\label{eq:asymmetric_lq_xhat} 
    \addtolength{\abovedisplayskip}{-1mm}
    \addtolength{\belowdisplayskip}{-1mm}   
    x_q = s (x_{int} - z)
\end{equation}

where $x$ is the input to be quantized, $x_{int}$ its integer version, $x_q$ the discrete representation on the same scale of $x$, $b$ the number of bits, $\nint{}$ the rounding operator, $s = \frac{\alpha_+ - \alpha_-}{2^b - 1}$  a "scale" that is function of the boundaries $\alpha_{\pm}$, and $z = - \alpha_- / s$ is the "zero point", that aligns $x_{int}$ to zero for $x = \alpha_-$. The $clamp$ function restricts $x_{int}$ range to $[0, 2^b -1]$. 

Symmetric LQ, which relies on a single boundary $\alpha$ and no zero point, is a less flexible representation but naturally maps zero inputs exactly to an integer in $x_{int}$ and, from a HW perspective, avoids the overhead incurred by performing MAC operations between values solely quantized with asymmetric LQ~\cite{wu2020integer}.

The quantization process establishes an inherent trade-off between clipping error and rounding error. The former arises by the approximation of the input values $x$ that lay outside the range $[\alpha_-,\alpha_+]$. The latter relates to the residual between each input $x$ and its rounded form $x_q$. For inputs within the clipping range, the residual is upper bounded by $s/2$, which increases as the $[\alpha_-,\alpha_+]$ range widens. 

Different quantizers can be chosen to determine the most appropriate $\alpha_{\pm}$ for a given weight or activation tensor. The simplest, herein named FIX, relies on a predetermined choice for $\alpha_{\pm}$ which remain fixed during QAT and inference. This is the fastest quantizer for QAT as it does not utilize any analysis of the distribution statistics nor gradient propagation on $\alpha_{\pm}$. However, it does not adapt to the input distribution and the choice of $\alpha_{\pm}$ may not be well motivated and sub-optimal. FIX is only applicable in specific cases, as discussed later.

A more common strategy, herein called MAX, is to set $\alpha_{\pm}$ to the extremes (min and max) of the distribution. MAX captures the whole range of values with no clipping error but suffers in the presence of outliers that increase the rounding error.

Statistics Aware Weight Binning (SAWB)~\cite{choi2019accurate} is a symmetric quantizer for weights that leverages the distribution statistics to identify the boundary $\alpha$ that most closely approximates the optimal $\alpha^*$ that minimizes the mean square error (a combination of clipping and rounding error). 

With Parameterized Clipping Activation (PACT)~\cite{choi2019accurate}, the boundaries $\alpha_{\pm}$ are independently learned during QAT. Learned quantizers require tracking the quantization operations, computing the corresponding gradients, and updating $\alpha_{\pm}$. Consequently, they are more computationally expensive compared to non-learned quantizers like FIX, MAX, or SAWB. The gap in computation time may be onerous in the QAT of large ASR models, due to the repeated calls to the activation quantizers.



\subsection{Model quantization}
\label{sec:models_quant}

\begin{figure}[t]
  \centering
  \includegraphics[width=\linewidth]{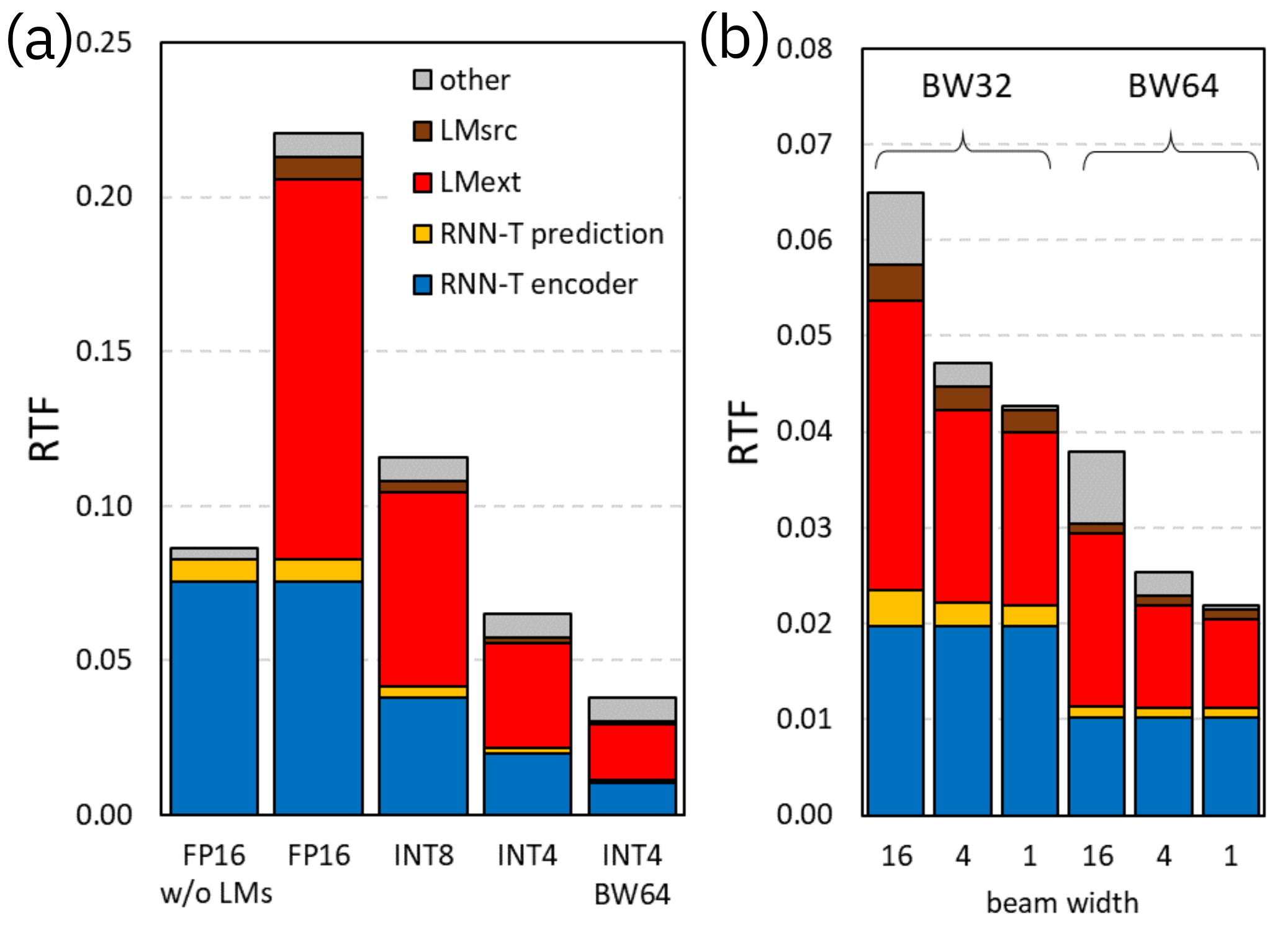}
  \caption{Real Time Factor (RTF) of quantized RNN-T decoding with density ratio LM fusion on simulated custom HW: (a) impact of reduced precision, at beam width 16 (BW64 = 64~Gbps bandwidth); (b) beam width-dependence of 4-bit model, at 32 and 64 Gbps. Legend applies to both figures.}
  \label{fig:runtime}
  \vspace{-4mm}
\end{figure}

Fig.~\ref{fig:arch} presents our main quantization scheme, as applied to an RNN-T model inclusive of LM fusion (LM\textsubscript{EXT} and LM\textsubscript{SRC}). Weights (without biases) and activations are quantized across the whole network. INT4 quantization is shown in blue, INT8 in red. The quantizers are indicated by the abbreviations within each cell (F = FIX, M = MAX, P = PACT, S = SAWB). For LSTM layers, 3 quantizers are used, for weights, inputs, and hidden states. The LSTM cell states are not quantized as they only participate in element-wise computations and their contribution to the total operations is minimal. In the linear layers, weights and the inputs are quantized. To limit HW overhead during MAC operations~\cite{wu2020integer}, symmetric quantizers are used for the weights across the whole network. Re-training the RNN-T with QAT is computationally expensive. Quantization of activations is critical because RNN models require quantizers at every time step. This is not the case for the weights, which can be quantized just once per utterance. Therefore, practical considerations drive the selection towards less expensive activation quantizers, provided that accuracy is not impacted.

With reference to Fig.~\ref{fig:arch}, all the bidirectional LSTM layers of the RNN-T acoustic encoder \textit{except the first} are quantized at 4 bits. We use SAWB to determine an optimal $\alpha$ boundary for the weights, as the rounding error introduced by the MAX quantizer at 4 bits is detrimental to accuracy. For inputs and hidden states we use the FIX quantizer in a shared configuration across time steps. In the LSTM layers, inputs and hidden states are the results of the computation taking place at the previous layer or time step, respectively, and in particular of the Hadamard product of two activation functions: $\tanh()\odot\sigma()$. Consequently, each tensor element is in [-1,1] (before dropout scaling). Therefore, we initialize the $\alpha_{\pm}$ boundaries of these FIX quantizers to $\pm1$. We also note that the upper boundary obtained with a MAX quantizer when utilized in this scenario typically falls in the interval [0.9,1], which further supports the selection of the chosen boundaries of the FIX quantizer.

The first layer of the network has major impact on the accuracy, while accounting for $<10\%$ of parameters and computations of the encoder. Therefore, we use 8-bit quantization on this layer, with a MAX quantizer for both weights and inputs. 8-bit MAX provides sufficient granularity to limit the rounding error in the presence of outliers, while not restricting the range to an arbitrary value. On the other hand, 8-bit FIX with $\alpha_{\pm} = \pm1$ remains appropriate for the hidden states. On the prediction side, we quantize the LSTM layer to 4 bits using SAWB for weights, and FIX for inputs ($\alpha_{\pm} = \pm1.25$) and hidden states ($\alpha_{\pm} = \pm1$). The remaining linear layers are quantized at 8 bits using MAX for both weights and activations. At 4 bits, they bring substantial degradation while accounting for a small fraction of parameters and computations ($<1\%$).

The re-training of the LM models is less computationally expensive and affords for more flexibility in the choice of quantizers. Therefore, we fully quantize both LM\textsubscript{EXT} and LM\textsubscript{SRC} at 4 bits, using SAWB for weights and PACT for inputs. With this configuration, we achieve a 7.6$\times$ reduction in model size compared to the FP32 model, with $94.9\%$ of parameters at 4-bits.

\subsection{Quantization Aware Training}
\label{sec:qat}

The RNN-T, LM\textsubscript{EXT}, and LM\textsubscript{SRC} are re-trained separately. The QAT process mirrors the non-quantized training, with a few important differences. First, we adjust the initial LR and LR schedule to ensure the model can properly learn to compensate for the error introduced by the quantization. Following~\cite{fasoli21lstm}, we lower the initial RNN-T LR to $4\cdot10^{-4}$ and decrease it linearly to a lower bound of $10^{-5}$. Then, we shorten the QAT to $<$15 epochs and train with batch size 768. For QAT of the LMs we retain the same LR schedule but lower the initial LR to $8\cdot10^{-4}$. 

\subsection{Inner vs. Outer quantization}
\label{sec:in_n_out}

It is common to conceptualize the quantization as "per layer", where the quantizers convert both the weights and the input tensors into their discretized form. In RNNs, the time dependence also requires activation quantization within layers, across time steps. One option is to quantize the inputs $x_t$ and the hidden states $s_{t-1}$ as they ``enter'' the RNN cell, before performing the MAC with the corresponding weights $W_{ih}$ and $W_{hh}$. We refer to this approach as ``Inner'' quantization (fig.~\ref{fig:arch} inset). The number of quantization calls throughout a network is $2*T*L$ for 2 input tensors, with T the number of time steps and L the number of layers (counted twice for bi-dir RNNs).

This approach is inefficient. For each cell, the same output tensor $s_t$ is provided to both the next time step and layer. Therefore, we quantize it once, at the cell output (fig.~\ref{fig:arch} inset). We call this process ``Outer'' quantization. Assuming $s_{t-1}$ is zero at $t=1$, the number of quantization calls that Outer quantization requires is reduced to $T + T*L$. The ratio between Outer and Inner quantizer calls is $(L+1)/(2*L)$, independent of the sequence length. Outer quantization reduces the number of calls to activation quantizers in the RNN-T encoder by 45\%.


\section{Experimental results}

\begin{table*}[t]
  \hspace*{-1cm}
  \centering
  \footnotesize
  \setlength{\tabcolsep}{3pt}
  \caption{Network acceleration (vs. FP16), compression ratio (vs. FP32), and WER at various quantization stages of RNN-T and LMs. Quantization is expressed as bits of weights/activations. Beam width (bm) is 16 unless otherwise stated. $\Delta$ is WER \textit{gap} against a model with equal RNN-T quantization and no LM fusion. All networks are initialized with FP32 pre-trained weights}
  \label{tab:wer}
    \begin{tabular}{lllrrr|rrrlrr|rrrrl|rrrl}
      \multicolumn{6}{c|}{}                   & \multicolumn{6}{c|}{\textbf{Hub5 2000}} & 
      \multicolumn{5}{c|}{\textbf{Hub5 2001}} & \multicolumn{4}{c}{\textbf{RT 2003}} \\
      \midrule
      \multicolumn{1}{c}{\textbf{RNN-T}} & 
      \multicolumn{1}{c}{\textbf{LM\textsubscript{EXT}}} & 
      \multicolumn{1}{c}{\textbf{LM\textsubscript{SRC}}} & 
      \multicolumn{1}{p{0.7cm}}{\scriptsize{\textbf{accel.}} \tiny{\textbf{(v. FP16)}}} & 
      \multicolumn{1}{p{0.5cm}}{\scriptsize{\textbf{size (MB)}}} & 
      \multicolumn{1}{p{0.7cm}|}{\scriptsize{\textbf{compr.}}  \tiny{\textbf{(v. FP32)}}} &
      \multicolumn{1}{c}{\textbf{SWB}} &  
      \multicolumn{1}{c}{\textbf{CH}} & 
      \multicolumn{1}{c}{\textbf{avg}} & 
      \multicolumn{1}{c}{\textbf{$\Delta$}} & 
      \multicolumn{1}{p{0.5cm}}{\textbf{avg bm8}} & 
      \multicolumn{1}{p{0.5cm}|}{\textbf{avg bm4}} & 
      \multicolumn{1}{c}{\textbf{SWB}} &  
      \multicolumn{1}{c}{\textbf{S2P3}} & 
      \multicolumn{1}{c}{\textbf{S2P4}} & 
      \multicolumn{1}{c}{\textbf{avg}} & 
      \multicolumn{1}{c|}{\textbf{$\Delta$}} &  
      \multicolumn{1}{c}{\textbf{SWB}} &  
      \multicolumn{1}{c}{\textbf{FSH}} & 
      \multicolumn{1}{c}{\textbf{avg}} & 
      \multicolumn{1}{c}{\textbf{$\Delta$}} \\
      \midrule
      32/32 & -       & -       &     & 223.3 &     & 8.1 & 15.5 & 11.8 &      & 11.8 & 11.9 & 8.5 & 11.7 & 15.9 & 12.1 &   & 18.5 & 11.8 & 15.3 & \\
      32/32 & 32/32   & 32/32   &     & 432.5 &     & 6.3 & 13.1 & 9.8 & -2.0 & 10.0 & 10.6 & 7.1 & 9.4 & 13.6 & 10.1 & -2.0 & 15.4 & 9.5 & 12.5 & -2.8 \\
      \midrule
      4/4   & -       & -       & 3.4 &  30.9 &  7.2 & 8.4 & 16.4 & 12.4 & & 12.6 & 12.6 & 8.9 & 12.3 & 16.7 & 12.8 & & 19.9 & 12.9 & 16.5 & \\
      4/4   & 32/32   & 32/32   & 1.4 & 240.1 &  1.8 & 6.8 & 14.3 & 10.6 & -1.8 & 10.8 & 11.2 & 7.6 & 10.2 & 14.2 & 10.8 & -2.0 & 17.0 & 10.5 & 13.8 & -2.7 \\
      4/4   & 8/8 & 32/32   & 2.2 &  90.9 &  4.8 & 6.8 & 14.3 & 10.6 & -1.8 & 10.8 & 11.3 & 7.6 & 10.5 & 14.3 & 10.9 & -1.9 & 17.1 & 10.6 & 14.0 & -2.5 \\
      4/4   & 4/4 & 32/32   & 3.1 &  66.0 &  6.6 & 6.9 & 14.5 & 10.7 & -1.7 & 10.9 & 11.4 & 8.0 & 10.7 & 14.8 & 11.2 & -1.6 & 17.6 & 10.9 & 14.4 & -2.1 \\
      4/4   & 4/4 & 4/4 & 3.4 &  57.2 &  7.6 & 7.0 & 14.4 & 10.7 & -1.7 & 11.0 & 11.3 & 8.1 & 10.6 & 14.9 & 11.3 & -1.5 & 17.4 & 10.9 & 14.3 & -2.2 \\
      \midrule    
      2/4   & -       & -       &     &  18.3 & 12.2 & 9.0 & 17.4 & 13.2 & & 13.3 & 13.3 & 9.6 & 13.1 & 17.5 & 13.5 &  & 21.0 & 13.7 & 17.5 & \\
      2/4   & 4/4 & 4/4&     &  44.6 & 9.7 & 7.4 & 15.2 & 11.4 & -1.8 & 11.5 & 11.9 & 9.8 & 12.5 & 17.1 & 13.2 & -0.3 & 19.7 & 12.8 & 16.3 & -1.2
    \end{tabular}
  \hspace*{-1cm}
  \vspace{-1mm}
\end{table*}

\subsection{Runtime performance and acceleration}
\label{sec:rtf}

We perform a comparative evaluation of the runtime performance of the RNN-T with LM fusion. Following~\cite{Venkataramani2019,venkataramani2019memory,fasoli21lstm}, we simulate the workload on a custom accelerator consisting of a coprocessor and attached CPU~\cite{Agrawal2021}. The compute-heavy data-parallel operations are executed on the coprocessor, while control-heavy operations such as hypothesis sorting are mapped to the CPU. We evaluate accelerations against an FP16 baseline (instead of FP32) due to restrictions in our HW simulator.

Fig.~\ref{fig:runtime} compares the estimated RTF at different precisions, for a single 152-frame sequence with beam width 16. In the absence of additional LMs, the FP16 inference workload is dominated by the acoustic encoder, which accounts for 87\% of the RTF (fig.~\ref{fig:runtime}(a)). Smaller beam widths would only improve the decoding portion. Conversely, the model with LM fusion shows a strong contribution of the decoding loop, with the large LM\textsubscript{EXT} accounting for 56\% of the total RTF.

The quantization of the RNN-T with LM fusion brings about remarkable speedup in both the encoder and the decoder. RTF drops from 0.221 at FP16, to 0.116 on a full INT8 model, and further down to 0.065 with full INT4 quantization, improving the RTF of the original RNN-T without extra LMs (fig.~\ref{fig:runtime}(a)). At 4 bits, we estimate an acceleration of 3.8$\times$ for the encoder, 3.6$\times$ for the decoder (inclusive of prediction and LMs), and 3.4$\times$ for the E2E model. The non-accelerable operations account for 12\% of the total 4-bit workload. Our simulations also reveal sub-optimal coprocessor occupancy, indicating a CPU-coprocessor communication bottleneck which would be mitigated by a higher bandwidth (BW) connection. The last column of fig.~\ref{fig:runtime}(a) shows the RTF when modeling a BW increase, from 32 to 64 Gbps. The higher BW is highly beneficial, yielding a 6.8$\times$ acceleration of the decoding loop, 5.8$\times$ for the whole network and an RTF of 0.038.

Due to the BW bottleneck, the computation favors large beam widths. As shown in fig.~\ref{fig:runtime}(b), the decoding RTF decreases sub-linearly at small beams, primarily as a result of the increased reuse of cached hypotheses. More remarkable is the improvement in non-accelerable operations runtime, which becomes negligible at beam = 1 (greedy decoding). The higher BW improves RTF across all beam widths, allowing for a more considerable reduction in the decoder contribution, down from 0.038 at beam 16 to 0.022 at beam 1. However, as discussed in section~\ref{sec:accuracy}, lower beam widths reduce transcription accuracy. 

\subsection{Accuracy and compression}
\label{sec:accuracy}

Table~\ref{tab:wer} shows the WER results at different stages of the quantization. At beam width 16, the fusion process improves the non-quantized model WER by 2.0\% while almost doubling the model size. As discussed in section~\ref{sec:rtf}, LM fusion also increases inference runtime dramatically.  

INT4 quantization of the RNN-T alone (following the scheme shown in fig.~\ref{fig:arch} but without LM\textsubscript{EXT} and LM\textsubscript{SRC}) shows a small WER degradation across datasets (0.6\% on average for Hub5 2000) with a compression factor of 7.2$\times$ (vs. FP32) and a 3.4$\times$ acceleration (vs. FP16). Although implementation of the Density Ratio LM Fusion approach shows a remarkable WER improvement on the INT4 quantized RNN-T, it is apparent that it also calls for quantization of LM\textsubscript{EXT} and LM\textsubscript{SRC} to retain the gains in model size and acceleration.

We initialize the INT8 and INT4 LM\textsubscript{EXT} with pre-trained FP32 weights, which we found helpful to minimize WER degradation. The 4-bit LM\textsubscript{EXT} quantized as per Fig.~\ref{fig:arch}, achieves comparable WER to the FP32 counterpart across benchmarks. To further boost compression and acceleration, we use an equivalent strategy to quantize the LM\textsubscript{SRC} at 4 bits, with no change in performance. With both LMs fully quantized at 4 bits, we obtain a 1.7\% WER improvement on Hub5 2000 compared to the model without LM fusion, while benefiting from a close-to-ideal 7.6$\times$ compression ratio of the full model and a 3.4$\times$ acceleration.

In the beam width comparison, the INT4 model with LM fusion achieves at beam width 16 better WER than the full precision model without LMs at any beam widths, while providing a substantial reduction in model size and comparable RTF (as discussed in sec.~\ref{sec:rtf}).

More aggressive quantization of the RNN-T weights (INT2) shows increased but limited WER degradation in the model without LMs, $\leq$1\% across benchmarks compared to INT4 weights. On Hub5 2000, LM fusion has a similar impact on WER as in the 4-bit scenario ($\Delta$=-1.8\%). Here, we observe a trade off between WER, degraded from 10.7\% to 11.4\%, and compression ratio, improved from 7.6$\times$ to 9.7$\times$. The favorable impact of LM fusion on other benchmarks is less pronounced, showing that the error introduced by the 2-bit representation of the RNN-T weights can negatively affect the whole decoding process.

\section{Conclusions}

We presented a QAT approach at 4 bits which enables model compression and inference acceleration of RNN-T networks combined with external LM fusion methods with minimal WER degradation. We showed that an appropriate choice of quantizers that leverages the inherent boundaries of the LSTM activation functions and the connections between LSTM cells limits QAT model retraining time while maintaining excellent WER performance. The 4-bit quantized model with LM fusion improves the WER of the full precision model without LMs, while achieving comparable RTF (and vastly improved RTF compared to the full precision model with LMs) making it a viable option for real-time inference workloads.


\bibliographystyle{IEEEtran}

\bibliography{references}

\end{document}